\documentclass[conference, a4paper]{APSIPA2025} 
\usepackage{amsmath,amssymb,amsfonts}
\usepackage{algorithmic}
\usepackage{graphicx}
\usepackage{textcomp}
\usepackage{xcolor}
\usepackage{multirow}
\usepackage{caption}
\usepackage{float}
\usepackage[backend=biber, style=ieee, isbn=false, doi=false, url=false]{biblatex}
\usepackage[draft]{hyperref}

\usepackage{orcidlink}

\def\BibTeX{{\rm B\kern-.05em{\sc i\kern-.025em b}\kern-.08em
    T\kern-.1667em\lower.7ex\hbox{E}\kern-.125emX}}
\usepackage{geometry}
\usepackage{fancyhdr}

\usepackage{geometry}
\begin{document}

\title{\huge Cancelable Biometric Template Protection Based on\\ Multi-Instance Fusion: A Contralateral Iris Approach}

\author{
\IEEEauthorblockN{Jittarin Chaivong\IEEEauthorrefmark{1} %\orcidlink{0009-0003-7929-4485}
Nicha Vikromrotjananan\IEEEauthorrefmark{1}
%\orcidlink{0009-0004-0526-7932},
Teekatat Piriyapittaya\IEEEauthorrefmark{1} %\orcidlink{0009-0009-2258-4127}
\\Waree Kongprawechnon\IEEEauthorrefmark{1}
%\orcidlink{0000-0002-2057-6352},\ 
Suradej Duangpummet\IEEEauthorrefmark{2}
%\orcidlink{0000-0001-7145-880X}
}
\IEEEauthorblockA{\IEEEauthorrefmark{1}\textit{Sirindhorn International Institute of Technology}, Thammasat University, Thailand\\
E-mail: \{m6822040348, m6822040363, 6722781844\}@g.siit.tu.ac.th, waree@siit.tu.ac.th
}
\IEEEauthorblockA{\IEEEauthorrefmark{2}National Science and Technology Development Agency, Thailand\\
E-mail: suradej.dua@nectec.or.th
}
}

\maketitle

% --- Abstract & Keywords ---
\begin{abstract}

Biometric templates are vulnerable to theft if stored without protection. Unlike passwords, a compromised iris cannot be reissued. Although existing cancelable biometric schemes address this problem, most still require an external key or token, introducing an additional attack surface. This paper proposes a cancelable contralateral iris template protection scheme that eliminates the need for a separate token or stored secret, satisfying the three requirements of ISO/IEC 24745: irreversibility, unlinkability, and confidentiality. The method fuses three enrollment samples per eye using Majority Vote Fusion to produce a stable template, and applies a salt-based bit permutation derived from the subject's enrollment ID. Combining the left- and right-permuted templates via a bitwise XOR produces a single Protected Fused Template. Since left and right iris patterns are statistically independent, fusing contralateral irises improves the accuracy of the system. An attacker must possess both iris codes and both salts to recover any useful information, yielding a larger effective key space than single-iris schemes. Experiments on three datasets, CASIA-IrisV4-Interval, CASIA-IrisV2 (two devices), and CASIA-Iris-Thousand, yield EERs of $0.36$\%, $4.88$\%, $10.80$\%, and $3.35$\%, respectively; the highest value reflects the more challenging cross-device scenario. These results demonstrate that our contralateral approach outperforms unprotected baselines while remaining competitive with state-of-the-art cancelable methods. %In addition, an ablation study confirms the benefit on both recognition performance and security. To the best of our knowledge, this is the first scheme to combine tokenless multi-instance fusion and contralateral binding under ISO/IEC 24745.
\end{abstract}

% Privacy Preserving
% --- Main Sections ---

\section{Introduction}

Iris recognition is considered one of the most accurate biometric modalities and is widely deployed in high-security systems \cite{b1}. Iris patterns are highly distinctive and stable due to epigenetic randomness during fetal development, resulting in iris codes that exhibit high entropy and near-random bit distributions \cite{b13,b14}. These characteristics provide stable discriminative capability for individual identification and contribute to the high accuracy of iris-based biometric systems. Despite these advantages, iris template compromise remains a critical security challenge. Because biometric traits are permanent, a leaked iris template cannot be resolved simply by re-enrollment. This limitation exposes authentication systems to long-term security risks. 

Consequently, several cancelable biometric techniques have been proposed to transform biometric templates into revocable representations while preserving recognition performance \cite{b2,b3,b4,b5}. BioHashing-based methods \cite{b5} generate protected templates using external tokens or salts. Lee et al.~\cite{b7} proposed an alignment-robust cancelable biometric, while Zheng et al.~\cite{b10} introduced a slicing-based method that generates revocable iris templates through transformation, allowing compromised templates to be replaced without requiring new biometric samples. However, these methods rely on external tokens, introducing a secondary attack surface; if the token is compromised, system security is significantly weakened. Hence, Ouda et al.~\cite{b12} proposed a tokenless scheme that eliminates the need for external secrets. These approaches effectively improve template security and revocability; however, they are primarily designed for single-instance iris systems and do not exploit the rich information available from both irises.

Meanwhile, there is limited research on bilateral cancelable iris templates. Rafiq and Selwal~\cite{b6} proposed a scheme that combines left and right iris templates using a block-XOR operation. While this method enhances security and recognition robustness, its fusion strategy increases computational complexity. % and may not fully address the alignment variations required for robust cancelable systems. 
Later, Arepalli and Boobalan proposed a multi-instance framework, termed the Compressed Transformed Fused template~\cite{b8}. Although their method achieves an average processing time of $0.1$ seconds and an Equal Error Rate (EER) of $0.15$\%, it focuses only on compression and computational speed rather than optimal security. Furthermore, the aforementioned methods remain restricted to either cancelable mechanisms or multi-instance frameworks, rather than integrating both. Because the field still relies on single-iris templates, most systems fail to leverage the potential of multi-instance fusion to simultaneously enhance recognition accuracy and template security.

Therefore, this paper introduces a tokenless, multi-instance cancelable iris recognition framework that integrates bilateral information from both the left and right irises. Since the left and right iris patterns of an individual are statistically independent \cite{b13,b14}, bilateral irises serve as complementary sources of biometric information, enabling mutual verification \cite{Jain&Ross2004}. We employ contralateral XOR fusion to strengthen template protection, along with Majority Vote Fusion to create a stable enrollment template, ROI-based iris code slicing to improve recognition accuracy, and salt-based bit permutation for template randomization. Ultimately, this scheme fully satisfies the requirements of ISO/IEC 24745~\cite{b16}.%: irreversibility, unlinkability, and confidentiality.

\begin{figure*}[ht]
    \centering
    \includegraphics[width=0.95\textwidth]{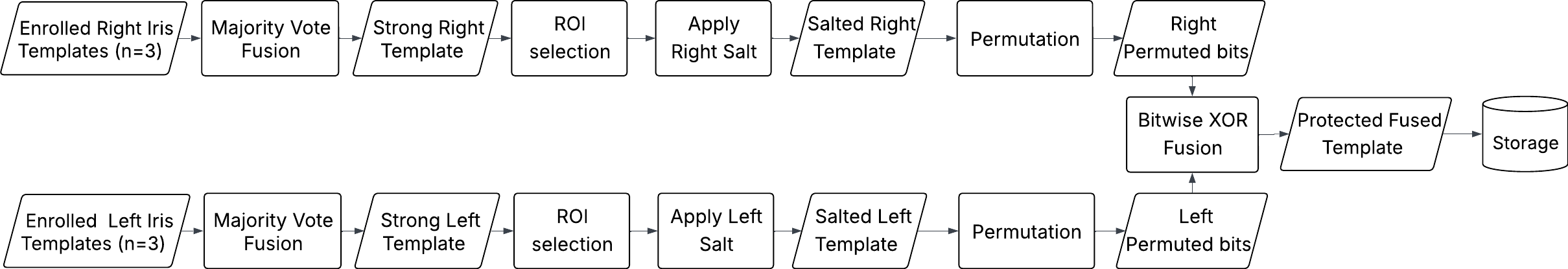}
    \caption{Enrollment phase: Three iris images per eye are fused via Majority Vote to produce a stable template, which undergoes ROI selection, salt-based bit permutation, and contralateral XOR binding to yield the stored Protected Fused Template (PFT).}
    \label{fig:enrollment}
\end{figure*}

\begin{figure*}[htbp]
    \centerline{\includegraphics[width=0.95\textwidth]{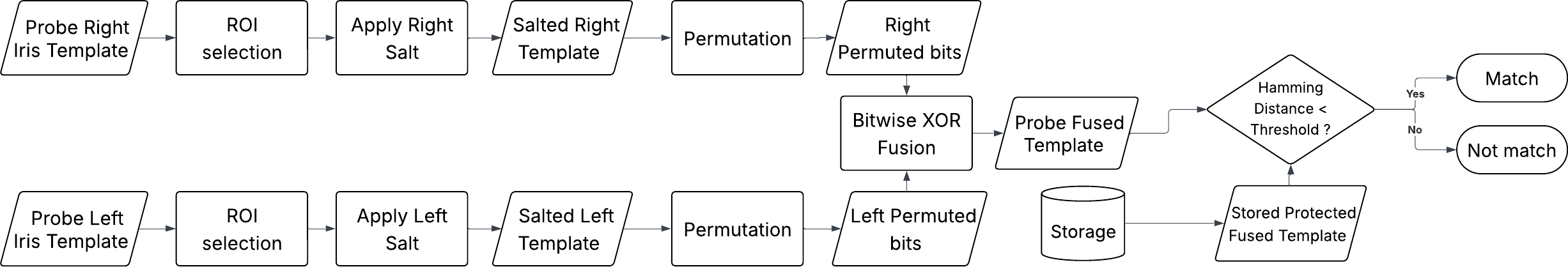}}
    \caption{Authentication phase: A single probe image per eye undergoes the same ROI selection and salt-based permutation as in enrollment. The left and right permuted codes are XOR-fused and compared against the stored PFT via Hamming distance.}
    \label{fig:authentication}
\end{figure*}

\section{Proposed Method}
\label{sec:proposed}

This section describes the proposed Cancelable Contralateral Fusion Scheme, which consists of preprocessing, feature extraction, region-of-interest selection, and the subsequent enrollment and authentication phases.

\subsection{Preprocessing and Feature Extraction}
The preprocessing and feature extraction are built based on an open-source library \cite{b17}, consisting of three steps:

\subsubsection{Segmentation and Normalization}
The iris, pupil, and sclera boundaries are identified through segmentation, yielding an inner circle (pupil--iris boundary) and an outer circle (iris--sclera boundary). The iris texture is then projected from Cartesian to polar coordinates \cite{b17}, producing a fixed-size two-dimensional rectangular array invariant to pupil dilation.

\subsubsection{Feature Extraction}
\label{subsubsec:feature_extraction}
Following normalization, two scales of two-dimensional Gabor wavelet filtering are applied to extract a binary iris code following Daugman's method \cite{b1}. The phase response is quantized into binary values, stored in an array of shape $[2, 16, 256, 2]$, where the first index represents two sizes of filter which are high-frequency and low-frequency filters, the second index represents each ring of the iris, the third index represents angular positions, and the last index represents the real and imaginary parts of the Gabor filter response \cite{b17}.

\subsection{Region of Interest Selection}
The inner rows of the iris code (rows 0–1, nearest the pupil boundary) are susceptible to artifacts from pupil dilation, while the outer rows (rows 7–15, near the iris–sclera boundary) are more frequently occluded by eyelids and eyelashes. Rows 2 through 6 capture the most stable and discriminative iris texture. Hence, we select these five rows as the Region of Interest (ROI) to distinguish between genuine and impostor samples. The feature vector consists of 5,120 bits per iris ($2~\text{filter sizes} \times 5~\text{rows} \times 256~\text{angular positions} \times 2~\text{phases (real and imaginary parts)}$). The justification is reported using the sensitivity metric (Section~\ref{subsec:d-prime}), and analysis in Section~\ref{subsec:ROI analysis}.

\subsection{Cancelable Contralateral Fusion Scheme}
The proposed method combines both left and right iris codes into a single Protected Fused Template (PFT). Because contralateral iris patterns are statistically independent even in the same person or twin \cite{b13,b14}, binding both canceled iris codes requires an attacker to possess both sides of the codes and their associated salts to recover any meaningful information.% This provides a stronger security guarantee than cancelable single-iris schemes \cite{b6,b7,b8}.
The method operates in two phases:

\subsubsection{Enrollment Phase}
As shown in Fig.~\ref{fig:enrollment}, three enrolled iris images per eye side are captured and fused by Majority Vote to establish the stable template for each eye. An eye-side-specific salt is applied to the stable template, permuting the bit positions to get the permuted bits. A bitwise exclusive-OR (XOR) operation is applied to the left and right sides of the permuted bits to create the PFT. Finally, only the PFT is stored, and all enrolled iris images are discarded.

\subsubsection{Authentication Phase}
As shown in Fig.~\ref{fig:authentication}, a new pair of iris images is captured. The same salts derived from the enrollment ID for each side are applied to the new template, permuting the bit locations to get the permuted bits. Left and right side permuted bits are applied to the XOR operation to get the probed template. Finally, the PFT is compared with the stored PFT by using the Hamming Distance score. If the score is below the threshold, the system will accept the authentication.

\section{Security Analysis}
\label{sec:security}

This section analyzes how the proposed method satisfies the three security requirements of ISO/IEC 24745 \cite{b16}: irreversibility, unlinkability, and confidentiality.

\subsection{Irreversibility}
The Protected Fused Template (PFT) is defined as:
\begin{equation}
\text{PFT} = \pi_R(T_R \oplus S_R) \oplus \pi_L(T_L \oplus S_L),
\end{equation}
where $T_R, T_L \in \{0,1\}^{5120}$ are sliced iris templates (one per eye), $S_R, S_L$ are per-eye 5,120-bit pseudorandom salts, $\oplus$ is the XOR operation, and $\pi_R, \pi_L$ are bit permutations that are derived from the subject's enrollment ID. Only the PFT is stored; the original iris captures and sliced templates $T_R, T_L$ are all discarded. In addition, the enrollment phase utilizes a majority vote from iris templates.
The irreversibility of this method consists of three layers:

\subsubsection{Layer 1: ROI selection}
\label{subsec:ROI analysis}
Only rows 2--6 of the iris code, the 5,120-bit ROI, are utilized, and rows 0--1 (inner zone) and 7--15 (outer zone) are permanently discarded. The full iris code cannot be recovered. Although the attacker can reverse all transformations, they can recover only ROI sliced template.

\subsubsection{Layer 2: Majority Vote Fusion}
During the enrollment phase, the sliced iris templates ($T_R$ and $T_L$) are not derived from a single iris capture. It is a per-bit majority vote over $n=3$ independent captures. This voted template is a many-to-one function of the enrollment images. This does not reveal any information about any individual capture, and the original enrollment images cannot be reconstructed from them.

\subsubsection{Layer 3: Salt permutation and contralateral XOR}
To recover $T_R$ from the PFT, an attacker must know $S_R$, $S_L$, $\pi_L$, and $T_L$. Even if an attacker already knows $T_R$, the XOR with $\pi_L(T_L \oplus S_L)$ still hides $T_L$ completely. Since each salt is a 5,120-bit pseudorandom string derived from the enrollment ID, reconstructing the salt requires knowing the enrollment ID. The method assumes that the enrollment ID is kept confidential and not stored in the same database as the PFT.

\subsection{Unlinkability}
A pair of $(S_R, S_L)$ is issued independently for each database. Suppose the same user is enrolled in two databases with salt pairs $(S_R^{(1)}, S_L^{(1)})$ and $(S_R^{(2)}, S_L^{(2)})$, producing $\text{PFT}_1$ and $\text{PFT}_2$. Because of the salt differentiation, each side of the permuted templates is statistically independent. Under this assumption, the expected Hamming distance should be approximately 0.50 \cite{b1}:
\begin{equation}
\text{HD}(\text{PFT}_1,\, \text{PFT}_2) \approx 0.50.
\end{equation}
This guarantees that the same user's templates across different databases are statistically indistinguishable, satisfying the unlinkability requirement.

\subsection{Confidentiality}
The proposed method satisfies confidentiality because only the PFT is stored in the database. All original iris and sliced templates, $T_R$ and $T_L$, are permanently discarded after enrollment. A leaked PFT reveals no raw biometric data. To recover any iris template from the PFT, one needs to know both salts and the other side's iris code, as derived above. The enrollment ID is assumed to be kept confidential and not in the same location as the PFT. Thus, an attacker who obtains only the stored PFT cannot gain any usable biometric information, satisfying the confidentiality requirement of ISO/IEC 24745~\cite{b16}.

\section{Results and Discussion}
\label{sec:results}
%The following subsections report the experimental setup, results, and security evaluation.

\subsection{Dataset}
Three datasets, CASIA-IrisV4-Interval, CASIA-IrisV2, and CASIA-Iris-Thousand, were used. The first dataset is the CASIA-IrisV4-Interval, containing 2,639 iris pictures with a resolution of $320 \times 280$ pixels from 249 subjects. 121 subjects, with a total of 2,621 images and more than three image pairs for both eyes. The second dataset, CASIA-IrisV2, contains 1,200 iris images per device, each at a resolution of $640 \times 480$ pixels, from 30 subjects. All subjects are used with separate devices. The first device uses 1,187 pictures, and the second uses 1,181 pictures. The last dataset is the CASIA-Iris-Thousand, which contains 20,000 iris images with a resolution of $640 \times 480$ pixels from 1,000 subjects. There are 997 subjects (18,483 images in total) with contralateral irises used in the proposed method.

For each enrolled subject, the remaining iris images not used in the three-shot enrollment are paired into genuine and impostor test inputs at a 1:4 genuine-to-impostor ratio. A genuine pair consists of one right iris and one left iris probe image from the same enrolled subject. The impostor pool is constructed from three categories of equal size:
\begin{enumerate}
    \item \textit{Cross-subject impostors:} both right iris and
    left iris probes are drawn from a different enrolled subject.
    \item \textit{Mixed-eye impostors (genuine R, impostor L):}
    the right iris probe belongs to the enrolled subject, while
    the left iris probe is drawn from a different enrolled subject.
    \item \textit{Mixed-eye impostors (genuine L, impostor R):}
    the symmetric case, where the left iris probe is genuine and
    the right iris probe is from a different subject.
\end{enumerate}
The second and third categories directly model the \emph{mixed-eye attack}, in which an attacker possessing one genuine iris attempts to authenticate by pairing it with a stolen or synthetic contralateral iris. These categories constitute two-thirds of the impostor pool, ensuring that the reported EERs reflect resistance to this attack scenario, not only conventional cross-subject impostors.

\subsection{Evaluation Metrics}
The evaluation metrics were used as follows:

\subsubsection{D-prime, $d'$}
\label{subsec:d-prime}
 The $d'$ is the sensitivity metric for distinguishing between genuine and impostor Hamming distance distributions.
\begin{equation}
d' = \frac{|\mu_G - \mu_I|}{\sqrt{\frac{1}{2}(\sigma_G^2 + \sigma_I^2)}},
\end{equation}
where:
\begin{itemize}
    \item $\mu_G, \sigma_G$: Mean and standard deviation of the Genuine (intra-class) Hamming distances.
    \item $\mu_I, \sigma_I$: Mean and standard deviation of the Imposter (inter-class) Hamming distances.
\end{itemize}

\subsubsection{Equal Error Rate (EER)}
The EER is the error rate at the operating point where the False Match Rate (FMR) equals the False Non-Match Rate (FNMR):
\begin{equation}
\text{EER} = \text{FMR}(\tau^*) = \text{FNMR}(\tau^*),
\end{equation}
where $\tau^*$ is the Hamming Distance threshold where both FMR and FNMR are equal, and
\begin{align}
\text{FMR}(\tau) &= \frac{|\{\text{impostor pairs} : \text{HD} < \tau\}|}{|\{\text{all impostor pairs}\}|}. \\
\text{FNMR}(\tau) &= \frac{|\{\text{genuine pairs} : \text{HD} \geq \tau\}|}{|\{\text{all genuine pairs}\}|}.
\end{align}
The probe is accepted as genuine when $\text{HD} < \tau$. The lower the EER, the better the separation between genuine and impostor.

\subsubsection{Detection Error Tradeoff (DET) Curve}
The DET curve plots FNMR against FMR across all decision thresholds $\tau$, providing a comprehensive view of the trade-off between false non-matches and false matches at every operating point. Unlike a single EER value, the DET curve reveals the system's full operating characteristics. A curve closer to the origin indicates better overall performance.

\subsubsection{Hamming Distance}
The Hamming Distance (HD) measures the bit-level dissimilarity between two binary iris templates $\mathbf{A}, \mathbf{B} \in \{0,1\}^N$:
\begin{equation}
\text{HD}(\mathbf{A}, \mathbf{B}) = \frac{1}{N}\sum_{i=1}^{N} A_i \oplus B_i,
\end{equation}
To compensate for rotational misalignment during capture, the minimum HD over small cyclic shifts $k$ is taken:
\begin{equation}
\text{HD}^*(\mathbf{A}, \mathbf{B}) = \min_{k}\; \text{HD}\!\left(\mathbf{A},\, \sigma^k(\mathbf{B})\right),
\end{equation}
where $\sigma^k(\cdot)$ denotes a cyclic shift by $k$ positions.

\subsection{ROI Selection}
To identify the most discriminative region between genuine and impostor, we compute the sensitivity index $d'$ per row across all datasets, as shown in Table~\ref{tab:dprime}. The inner zone (rows 0 and 1) shows significantly lower $d'$ than row 2, while the outer zone (rows 7 to 15) demonstrates the lowest $d'$ compared to other zones. Rows 2 to 6 (the middle zone) show the highest $d'$ across all datasets and are therefore selected as the ROI.

\begin{table}[htbp]
\caption{Per-row $d'$ separation between genuine and impostor Hamming distance distributions (rows 2--6 in bold are the selected ROI).}
\label{tab:dprime}
\centering
\begin{tabular}{ccccc}
\hline
\textbf{Row} & \textbf{CASIA-IrisV4} & \textbf{CASIA~2} & \textbf{CASIA~2} & \textbf{CASIA} \\
             & \textbf{Interval} & \textbf{Dev.~1} & \textbf{Dev.~2} & \textbf{Thousand} \\
\hline
\multicolumn{5}{l}{\textit{Inner zone}} \\
0  & 4.47 & 2.51 & 2.37 & 2.25 \\
1  & 5.30 & 2.97 & 2.85 & 3.02 \\
\hline
\multicolumn{5}{l}{\textit{Middle zone (selected ROI)}} \\
\textbf{2}  & \textbf{5.91} & \textbf{3.41} & \textbf{3.02} & \textbf{3.70} \\
\textbf{3}  & \textbf{5.91} & \textbf{3.47} & \textbf{2.97} & \textbf{3.79} \\
\textbf{4}  & \textbf{5.71} & \textbf{3.41} & \textbf{3.01} & \textbf{3.65} \\
\textbf{5}  & \textbf{5.45} & \textbf{3.46} & \textbf{2.86} & \textbf{3.48} \\
\textbf{6}  & \textbf{5.01} & \textbf{3.42} & \textbf{2.70} & \textbf{3.27} \\
\hline
\multicolumn{5}{l}{\textit{Outer zone}} \\
7  & 4.49 & 3.39 & 2.66 & 3.02 \\
8  & 4.05 & 3.24 & 2.55 & 2.72 \\
9  & 3.69 & 3.19 & 2.43 & 2.47 \\
10 & 3.34 & 2.96 & 2.35 & 2.27 \\
11 & 2.99 & 2.77 & 2.25 & 2.10 \\
12 & 2.67 & 2.62 & 2.14 & 1.96 \\
13 & 2.24 & 2.30 & 2.03 & 1.82 \\
14 & 1.84 & 2.09 & 1.89 & 1.70 \\
15 & 1.59 & 1.88 & 1.79 & 1.61 \\
\hline
\end{tabular}
\end{table}

\subsection{Ablation Study}
\begin{table}[H]
\centering
\caption{Ablation Study of the Proposed Cancelable Contralateral Iris Recognition (EER \%)}
\label{tab:ablation}
\begin{tabular}{lcccc}
\hline
\textbf{Ablation} & \textbf{CASIA} & \textbf{CASIA2} & \textbf{CASIA2} & \textbf{CASIA} \\
& \textbf{Interval} & \textbf{(Dev. 1)} & \textbf{(Dev. 2)} & \textbf{Thousand} \\
\hline
Proposed Method      & 0.36 & 4.88 & 10.80 & 3.35 \\
Without Permutation   & 0.18 & 3.22 &  9.71 & 2.38 \\
Without ROI selection & 0.54 & 4.60 &  9.41 & 4.33 \\
Without Majority Voting & 0.54 & 6.09 & 14.60 & 4.61 \\
Fixed Salt Key      & 0.36 & 4.88 & 10.80 & 3.35 \\
Without Salt          & 0.36 & 4.88 & 10.80 & 3.35 \\
\hline
\end{tabular}
\end{table}

As shown in Table~\ref{tab:ablation}, the ablation study was conducted under 1:4 genuine-to-impostor test inputs for observing the trade-off between accuracy 
and security. The permutation slightly increases EER but 
strengthens the security guarantee. The salt does not affect EER since it only permutes bit positions, but is 
essential for unlinkability. Majority Vote Fusion and ROI 
selection both contribute to recognition performance.

\subsection{DET Curve Evaluation}

\begin{figure}[b!]
    \centerline{\includegraphics[width=\columnwidth]{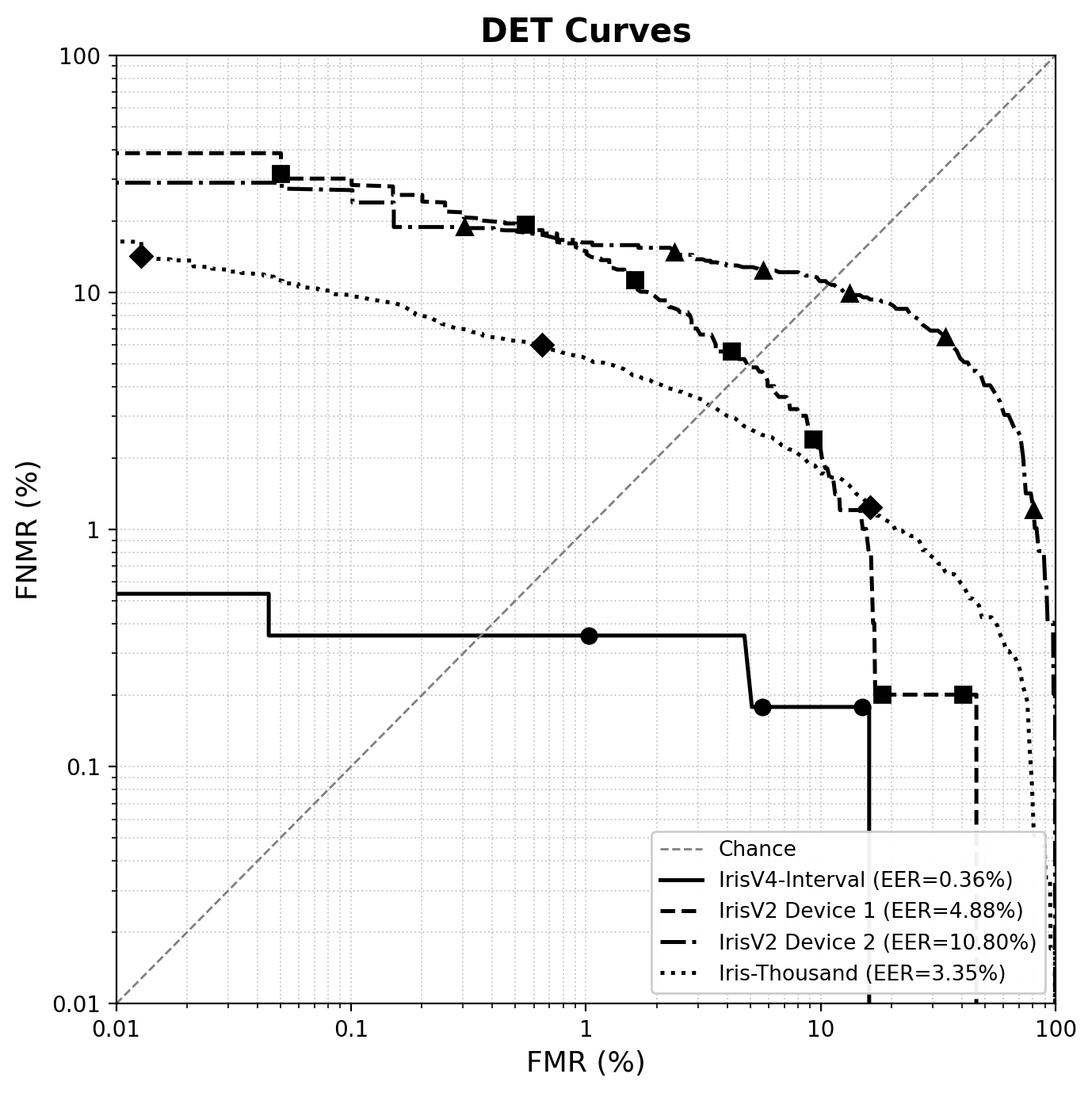}}
    \caption{DET curves (FNMR vs.\ FMR) of the proposed method on all three datasets (four experiments): CASIA-IrisV4-Interval, CASIA-IrisV2 Device~1, CASIA-IrisV2 Device~2, and CASIA-Iris-Thousand. Curves closer to the origin indicate better performance.}
    \label{fig:det}
\end{figure}

Fig.\,\ref{fig:det} demonstrates the DET curves of the proposed method across all three datasets. On CASIA-IrisV4-Interval, the curve approaches the origin closely, consistent with its $0.36$\% EER, showing strong separation between genuine and impostor. On CASIA-IrisV2 Device~1 and CASIA-Iris-Thousand, both curves remain close to the origin with their EERs of $4.88$\% and $3.35$\%, respectively. On CASIA-IrisV2 Device\,2, the curve is shifted toward the upper right due to the dataset's low separability with its $10.80$\% EER. The DET curves confirm consistent separability across all datasets at every operating point, not only at the EER.

\subsection{Comparison with Prior Works}

\begin{table*}[htbp]
\centering
\captionsetup{justification=centering}
\caption{EER comparison OF PROPOSED AND BASELINE METHODS.\\
The dash (``--'') indicates unreported values; $\checkmark$ and $\times$ denotes supported and unsupported feature, respectively.}
\label{tab:eer}
\centering
\begin{tabular}{lcccccc}
\hline
\textbf{Method} & \textbf{CASIA-IrisV4} & \textbf{CASIA~2} & \textbf{CASIA~2} & \textbf{CASIA} & \textbf{Multi-} & \textbf{Contra-} \\
                & \textbf{Interval}     & \textbf{Dev.~1}  & \textbf{Dev.~2}  & \textbf{Thousand} & \textbf{instance} & \textbf{lateral} \\
\hline
Quotient Filter~\cite{b8}   & 0.15  & --- & --- & --- & $\checkmark$ & $\times$ \\
Hybrid Ranking~\cite{b11}   & 0.16  & --- & --- & --- & $\times$     & $\times$ \\
Alignment-Robust~\cite{b7}  & 0.62  & --- & --- & 3.98 & $\times$   & $\times$ \\
Slicing~\cite{b10}          & 1.03  & --- & --- & --- & $\times$     & $\times$ \\
IrisSPT~\cite{b9}           & 1.70  & --- & --- & 4.43 & $\times$   & $\times$ \\
Tokenless CB~\cite{b12}     & 2.30  & --- & --- & --- & $\times$    & $\times$ \\
\hline
Unprotected (baseline)  & 3.06 & 11.27 & 16.63 & 11.45 & $\times$ & $\times$ \\
Unprotected (Rows 2--6)  & 2.68 & 12.68 & 16.84 & 10.12 & $\times$ & $\times$ \\
\hline
\textbf{Proposed Method} & \textbf{0.36} & \textbf{4.88} & \textbf{10.80} & \textbf{3.35} & $\checkmark$ & $\checkmark$ \\
\hline
\end{tabular}%
\end{table*}

Table~\ref{tab:eer} compares the proposed method against prior cancelable iris schemes and unprotected baselines. Prior works are evaluated only on CASIA-IrisV4-Interval (equivalent to CASIA-IrisV3-Interval: 2,639 images, 249 subjects); ``--'' indicates the unreported values. Because the proposed method fuses both irises, whereas all prior works operate on a single eye, contralateral binding provides an additional security layer not captured by single-iris EER alone. Across all datasets, the Proposed Method achieves the lowest EER compared to both unprotected baselines. The elevated EER of 10.80\% on CASIA-IrisV2 Device~2 reflects the dataset's low separability, even with an unprotected baseline that already reaches 16.63\%. Additionally, its per-row $d'$ values in Table~\ref{tab:dprime} are the lowest across all datasets. The proposed method still reduces the EER by approximately one-third relative to both unprotected baselines.

\subsection{Computation Time}

All experiments were conducted on an Apple MacBook Pro equipped with an Apple M5 chip (10-core CPU comprising 4 performance cores and 6 efficiency cores) and 16 GB unified memory, running macOS Tahoe 26.3. Each phase was measured and averaged over $100$ trials. Table~\ref{tab:time_stage} demonstrates the time each processing stage takes separately. %In addition, Table~\ref{tab:time_compare} demonstrates a comparison between the unprotected baselines and the proposed method.

\begin{table}[ht]
\caption{Average computation time (ms) per subject for each processing stage of the proposed method.}
\label{tab:time_stage}
\centering
\resizebox{\columnwidth}{!}{%
\begin{tabular}{llr}
\hline
\textbf{Phase} & \textbf{Stage} & \textbf{Time (ms)} \\
\hline
\multirow{4}{*}{Enrollment}
  & Feature extraction (6 images, open-iris) & 2,949.47 \\
  & Majority Vote Fusion (per eye side)      &     0.66 \\
  & Salt permutation (both eyes)             &     0.30 \\
  & XOR fusion $\rightarrow$ PFT             &     0.01 \\
\cline{2-3}
  & \textbf{Total enrollment}                & \textbf{2,950.44} \\
  & \textbf{Standard deviation}              & \textbf{50.65}    \\
\hline
\multirow{4}{*}{Authentication}
  & Feature extraction (2 images, open-iris) &   972.87  \\
  & Salt permutation (both eyes)             &     1.30  \\
  & XOR fusion $\rightarrow$ Probe PFT       &     0.49  \\
  & Hamming Distance comparison              &     1.05  \\
\cline{2-3}
  & \textbf{Total authentication}            & \textbf{975.71}    \\
  & \textbf{Standard deviation}              & \textbf{24.17}      \\
\hline
\end{tabular}%
}
\end{table}

\begin{table}[ht]
\caption{Computation time comparison across methods (ms per subject, averaged over 100 runs).}
\label{tab:time_compare}
\centering
\resizebox{\columnwidth}{!}{%
\begin{tabular}{lrr}
\hline
\textbf{Method} & \textbf{Enrollment (ms)} & \textbf{Authentication (ms)} \\
\hline
Unprotected (full iris)  & $2{,}974.12 \pm 74.59$  & $970.39 \pm 28.97$ \\
Unprotected (Rows 2--6)  & $2{,}953.26 \pm 93.55$  & $962.70 \pm 15.42$ \\
\textbf{Proposed Method} & $2{,}951.84 \pm 50.65$ & $978.26 \pm 24.17$ \\
\hline
\end{tabular}%
}
\end{table}

Table~\ref{tab:time_compare} demonstrates that the time difference between the proposed method and the unprotected baselines is negligible. The most time-consuming step across all methods is the feature extraction step from open-iris~\cite{b17}, which accounts for over $99$\% of the total time in both enrollment and authentication. The cancelable steps, such as Majority Vote Fusion, salt permutation, and XOR fusion, shown in Table~\ref{tab:time_stage}, total only about $0.97$~ms for enrollment, $1.79$~ms for authentication, and $1.05$~ms for Hamming Distance comparison. This demonstrates that the overhead introduced by the protection steps is negligible compared to feature extraction.

\subsection{Security Evaluation}

Each permutation $\pi_R$ and $\pi_L$ comes from a 5,120-bit salt that makes the search space per eye to be upper-bounded by $\log_2(5120!)$ bits. An attacker needs to know both $\pi_R$ and $\pi_L$ simultaneously without knowing either the salt or the raw iris code. Even if one iris code is fully known, for example, $T_R$ is leaked, the XOR with $\pi_L(T_L \oplus S_L)$ still completely hides $T_L$; thus, knowing one eye gives no advantage in recovering the other.
Beyond the mathematical security bound, as shown in Table~\ref{tab:eer}, the proposed method satisfies the three ISO/IEC 24745 requirements (irreversibility, unlinkability, and confidentiality), and additionally provides multi-instance fusion and contralateral binding. All prior methods \cite{b6,b7,b8,b9,b10,b11} operate on a single eye and do not use contralateral binding. Only \cite{b6} and \cite{b8} employ multi-instance fusion, yet they still rely on a single eye. No prior scheme simultaneously combines contralateral and multi-instance protection.

\section{Conclusion}
\label{sec:conclusion}
This paper presents a cancelable contralateral iris template protection scheme using Majority Vote Fusion ($n\,=\,3$ captures per eye), salt-based bit permutation, and contralateral XOR fusion to produce a single Protected Fused Template. To the best of our knowledge, this is the first scheme to combine tokenless multi-instance fusion and contralateral binding under ISO/IEC 24745. The proposed method was evaluated on three datasets (CASIA-IrisV4-Interval, CASIA-IrisV2 (device 1 and 2), and CASIA-Iris-Thousand), achieving EER of $0.36$\%, $4.88$\%, $10.80$\%, and $3.35$\%, respectively. The proposed method outperforms both the unprotected full-iris and ROI baselines on all datasets.

The proposed scheme satisfies three ISO/IEC 24745 requirements: irreversibility, confidentiality, and unlinkability. For irreversibility, reconstructing either iris code requires knowing both salts and the other eye's iris code. For confidentiality, original iris captures are permanently discarded after enrollment, and the stored PFT reveals no raw biometric data without knowing the enrollment ID. For unlinkability, the same subject enrolled in two databases with different enrollment IDs yields templates with an HD of approximately $0.50$, demonstrating complete statistical independence and preventing cross-database tracking.

However, it has a few limitations. Firstly, although the ROI selection (iris's rows~2--6) was optimized for CASIA datasets, it may not generalize to other datasets. Secondly, the scheme depends on a binary iris code produced by a 2D Gabor wavelet extractor, and the compatibility with other iris code extractors, such as deep learning-based feature extraction, has still not been evaluated. Lastly, the scheme is designed for scenarios in which both irises are available. Individuals with a single functional eye are excluded. Future work will address these limitations by evaluating the method with different sensors or in real-world deployment settings, exploring deep learning-based feature extraction, and extending it to multimodal biometric fusion.

%\newpage  %!!Use only if the page does not fit (camera-ready version)
\section*{Acknowledgment}
This research was financially supported by the Fundamental Fund under grant number FFB690024/0337, the National Research Council of Thailand (NRCT), the Thailand Advanced Institute of Science and Technology (TAIST), the National Science and Technology Development Agency (NSTDA), the Institute of Science Tokyo, the Sirindhorn International Institute of Technology (SIIT), Thammasat University (TU) under the TAIST–Science Tokyo Program.

% --- References ---

\printbibliography

\end{document}